\documentclass[journal]{IEEEtran}
\usepackage{amsmath,amsfonts}
\usepackage{algorithmic}
\usepackage{algorithm}
\usepackage{array}
\usepackage[caption=false,font=normalsize,labelfont=sf,textfont=sf]{subfig}
\usepackage{textcomp}
\usepackage{stfloats}
\usepackage{url}
\usepackage{verbatim}
\usepackage{graphicx}
\usepackage{cite}
\usepackage[utf8]{inputenc} 
\usepackage[T1]{fontenc}    
\usepackage{hyperref}       
\usepackage{url}            
\usepackage{booktabs}       
\usepackage{amsfonts}       
\usepackage{nicefrac}       
\usepackage{microtype}      
\usepackage{xcolor}         
\usepackage{graphicx}
\usepackage{booktabs}
\usepackage{multirow}
\usepackage{amsmath} 
\usepackage{adjustbox}
\usepackage{wrapfig}
\usepackage{floatflt}
\usepackage{xspace}

\makeatletter
\DeclareRobustCommand\onedot{\futurelet\@let@token\@onedot}
\def\@onedot{\ifx\@let@token.\else.\null\fi\xspace}

\def\eg{\emph{e.g}\onedot} 
\def\ie{\emph{i.e}\onedot}

\def\etal{\emph{et al}\onedot}
\makeatother

\usepackage[capitalize]{cleveref}
\crefname{section}{Sec.}{Secs.}
\Crefname{section}{Section}{Sections}
\Crefname{table}{Table}{Tables}
\crefname{table}{Tab.}{Tabs.}
\crefname{figure}{Fig.}{Fig.}

\begin{document}

\title{Towards Deeper Emotional Reflection: \\  Crafting Affective Image Filters \\ with Generative Priors}

\author{Peixuan Zhang$^\dagger$, Shuchen Weng$^\dagger$,  Jiajun Tang, Si Li*, and Boxin Shi,~\IEEEmembership{Senior Member, IEEE}
\IEEEcompsocitemizethanks{
\IEEEcompsocthanksitem{P. Zhang and S. Li are with the School of Artificial Intelligence, Beijing University of Posts and Telecommunications, Beijing 100876, China. Email: \{pxzhang, lisi\}@bupt.edu.cn.}
\IEEEcompsocthanksitem{S. Weng is with the Beijing Academy of Artificial Intelligence, Beijing 100083, China. Email: \{scweng\}@baai.ac.cn.}
\IEEEcompsocthanksitem{J. Tang and B. Shi are with the State Key Laboratory of Multimedia Information Processing and National Engineering Research Center of Visual Technology, School of Computer Science, Peking University, Beijing 100871, China. Email: \{jiajun.tang, shiboxin\}@pku.edu.cn.}
\IEEEcompsocthanksitem{$\dagger$ Equal contribution. $*$ Corresponding author.}
}
}

\maketitle

\begin{abstract}

 Social media platforms enable users to express emotions by posting text with accompanying images. 
 In this paper, we propose the Affective Image Filter (AIF) task, which aims to reflect visually-abstract emotions from text into visually-concrete images, thereby creating emotionally compelling results. 
 We first introduce the AIF dataset and the formulation of the AIF models. 
 Then, we present AIF-B as an initial attempt based on a multi-modal transformer architecture. 
 After that, we propose AIF-D as an extension of AIF-B towards deeper emotional reflection, effectively leveraging generative priors from pre-trained large-scale diffusion models. 
 Quantitative and qualitative experiments demonstrate that AIF models achieve superior performance for both content consistency and emotional fidelity compared to state-of-the-art methods.
 Extensive user study experiments demonstrate that AIF models are significantly more effective at evoking specific emotions.
 Based on the presented results, we comprehensively discuss the value and potential of AIF models.

\end{abstract}


\section{Introduction}

\IEEEPARstart{S}{ocial} media platforms (\eg, Facebook and X), as widely embraced communication tools, enable people to forge emotional connections with one another~\cite{KROSS202155}. 
By sending posts, users can share what they are seeing, talking, and thinking. 
Although texts in posts can express emotions straightforwardly, attached images may act as more powerful affective stimuli to attract the attention of their followers.

\begin{figure*}[t]
    \centering
    \includegraphics[width=\textwidth]{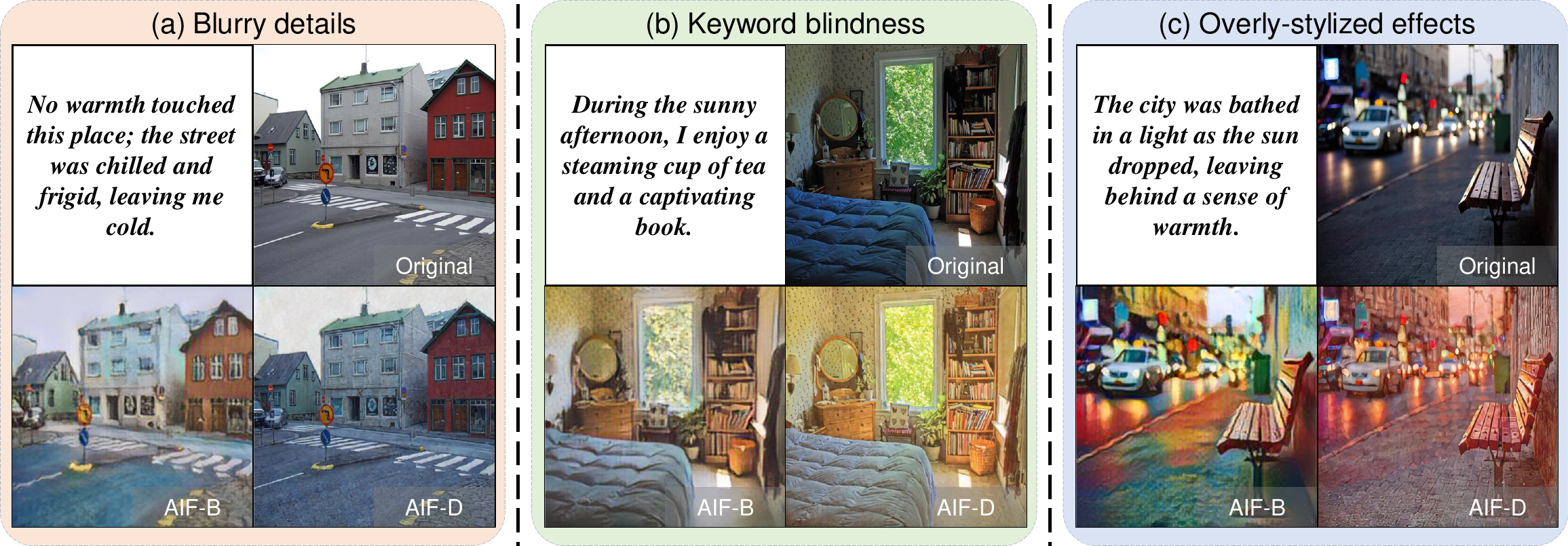}
    \caption{Towards deeper emotional reflection, we propose AIF-D to better meet the requirements of emotional fidelity and content consistency. Compared to previous AIF-B~\cite{aif}, AIF-D overcomes three major challenges: (a)~\textbf{Blurry details.} AIF-D preserves high-frequency details of the content image more effectively. (b)~\textbf{Keyword blindness.} AIF-D has a deeper emotional understanding of text descriptions that lack specific keywords. (c)~\textbf{Overly-stylized effects.} AIF-D reflects emotions with more appropriate artistic representation.} 
    \label{fig:teaser}
\end{figure*}

In an effort to help create unique and emotionally compelling images that stand out from the crowd, the task of Affective Image Filter (AIF)~\cite{aif} has emerged as an innovative research topic. As shown in~\cref{fig:teaser}, given a content image that depicts users' daily experience and a text description that reflects personal thoughts and feelings, an AIF model can reflect visually-abstract emotions from user-provided texts to visually-concrete images with appropriate colors and textures. Specifically, a qualified AIF model should exhibit both \textit{content consistency} and \textit{emotional fidelity}, which requires the model to effectively preserve the details of content images, accurately understand emotions in text descriptions, and appropriately visualize colors and textures with corresponding emotions.

The previous Basic AIF model~\cite{aif} (denoted as AIF-B) achieves the aforementioned objectives with a specifically collected AIF dataset. Since low-level and mid-level features (\eg, colors and styles) can effectively represent evoked emotions~\cite{machajdik2010affective, zhao2014affective}, the AIF dataset focuses on famous paintings across various aesthetic styles. Based on it, AIF-B takes the first step to concrete visualization of emotions, formulating the process as three parts: 
\textit{(i)} \textbf{Model architecture.} 
AIF-B is based on a multi-modal transformer architecture, which unifies content images and text descriptions into the token representation, enabling the model to learn their interdependency through subsequent transformer blocks. 
\textit{(ii)} \textbf{Emotional understanding.} 
To understand inherent emotional properties, AIF-B leverages the prior knowledge of the VAD dictionary~\cite{mohammad2018vad}, which encodes the valence, arousal, and dominance scores for emotional words. 
Additionally, AIF-B introduces an emotional distribution loss to capture high-dimensional emotional cues. 
\textit{(iii)} \textbf{Emotional visualization.}
To visualize the corresponding emotional expressions in synthesized images, AIF-B introduces a sentiment metric loss and an anchor-based sentiment loss to effectively reflect specific emotions. Content loss, style loss, GAN loss, and identity loss are further adopted to improve the aesthetic quality of synthesized images.

Although AIF-B has made remarkable progress in applying the corresponding filters, three major challenges still remain for this initial approach:
\textit{(i)} \textbf{Blurry details.} 
When applying AIF-B, a potential side effect is the removal of high-frequency details from objects, leading to a blurry effect (\cref{fig:teaser}~(a), synthesizing windows with blurred contours).
\textit{(ii)} \textbf{Keyword blindness.} To accurately evoke emotions, AIF-B may overly focus on specific keywords within text descriptions, potentially missing the broader context (\cref{fig:teaser}~(b), expressing feelings with less warmth and relaxation).
\textit{(iii)} \textbf{Overly-stylized effects.} When reflecting emotions to images, AIF-B may excessively apply artistic modifications, resulting in abrupt colors, exaggerated contrast, and distorted content (\cref{fig:teaser}~(c), visualizing emotions with over-saturated colors).

\IEEEpubidadjcol

In this paper, we extend our preliminary work AIF-B~\cite{aif} towards deeper emotional reflection. We design the Deeper AIF model (denoted as AIF-D) following the aforementioned formulation of AIF-B:
\textit{(i)} \textbf{Refined model architecture.}
We build AIF-D upon the pre-trained large-scale generative model~\cite{stablediffusion}, leveraging its rich generative priors to produce high-quality images. To address the blurry details caused by the downsampling block during image encoding, we design a content preservation module that extracts multi-scale spatial features. This module effectively preserves these details and maintains content consistency (\cref{fig:teaser} (a), the intricate details of the houses).
\textit{(ii)} \textbf{Nuanced emotional understanding.}
To avoid keyword blindness when handling complex emotional expressions, we leverage a Large Language Model (LLM)~\cite{llama2,gpt4v} and introduce Chain-of-Thought (CoT) prompting \cite{kojima2022large,li2024enhancing} to better reflect emotions from text descriptions. Additionally, we propose a voting ensemble mechanism that evaluates emotional distributions from multiple perspectives. By ensembling estimated emotional distributions, this mechanism enables accurate analysis of evoked emotions (\cref{fig:teaser}~(b), cozy atmosphere for the room). 
\textit{(iii)} \textbf{Advanced emotional visualization.}
To enable accurate visualization of emotions, we propose an emotional reflection strategy to effectively learn both emotional and artistic priors and present compelling artistic features.
After that, we redesign the aesthetic loss to balance artistic style and content consistency, ensuring that the synthesized images align more closely with the objectives of the AIF task. (\cref{fig:teaser}~(c), appropriate artistic representation in the street view). 

In summary, AIF-D improves content consistency and emotional fidelity, advancing the AIF models towards deeper emotional reflection by making the following contributions: 
\begin{itemize}
    \item We leverage the generative priors from the pre-trained generative model and propose a content preservation module to preserve high-frequency details. 
    \item We integrate an LLM and CoT prompting to improve emotional understanding. The voting ensemble mechanism is presented to accurately analyze evoked emotions. 
    \item We propose an emotional reflection strategy to enhance the image decoder. The aesthetic loss is further redesigned to balance artistic style and content consistency.
\end{itemize}

The remainder of this paper is organized as follows.
We begin with an introduction to the related work in~\cref{sec:related}, including text-guided image editing, visual-emotional analysis, and image style transfer.
We introduce the AIF dataset and AIF model formulation in~\cref{sec:setup}.
We then introduce AIF-B in~\cref{sec:aif-b}, followed by AIF-D in~\cref{sec:aif-d}, highlighting improvements made in model architecture, emotional understanding, and emotion visualization. 
In~\cref{sec:experiment}, we conduct both quantitative and qualitative experiments. 
In~\cref{sec:discussion}, we explore applications, failure cases, robustness testing, and differences from general editing methods. 
Finally, the paper is concluded in~\cref{sec:conclusion}.

\section{Related Work}
\label{sec:related}

\subsection{Text-guided Image Editing.}
As a user-friendly option, text descriptions have been widely used in image editing tasks, \eg, enabling modifications across color~\cite{lcoins, lcad}, texture~\cite{clva, aif}, and high-level semantic~\cite{flexit, tumanyan2023plug}. 
Initial research efforts are directed towards developing effective text injection modules~\cite{manigan, repaint, tdanet}.
The emergence of language-image pre-trained models~\cite{clip} further expands the scope of image editing, allowing for a broader, open-vocabulary approach~\cite{glide, diffusionclip, clipstyler}.
Recently, diffusion models~\cite{ddim, ddpm} have demonstrated remarkable improvement in text-to-image generation task~\cite{dalle2, imagen, stablediffusion}, inspiring researchers to leverage the powerful generative capabilities to advanced image editing applications, \eg, a fine-tuning model for multi-conditional image editing~\cite{controlnet, T2Iadapter, AIEdiT}, personalizing generative models through image inversion~\cite{dreambooth, multiconcept}, and developing advanced sampling strategies to steer generation results~\cite{prompt2prompt, pix2pixzero}. 
Leveraging rich generative priors from pre-trained models, AIF models could present rich details and create visually compelling synthesizing results.

\subsection{Visual Emotional Analysis.}
The visual emotional analysis holds a crucial position in computer vision~\cite{zhao2021affective}. 
In the early years, relevant efforts~\cite{machajdik2010affective} focus on utilizing hand-crafted features for emotional recognition in images.  
Various low-level features are adopted by numerous studies~\cite{zhao2014exploring, zhao2014affective} to represent emotional cues, \eg, color, shape, and texture. The potential of mid-level features in emotional recognition is subsequently identified~\cite{yang2018weakly, yao2020adaptive, rao2020learning}, providing a more intuitive human understanding than low-level features.
The emergence of deep learning has revolutionized this domain, researchers~\cite{yang2018visual, wang2022ease, yang2023context, wang2021deep} extract high-level features to improve performance in image-based emotional recognition by focusing on content-rich information, \eg, facial expressions. 
Following them, multiple emotional features are integrated for a comprehensive understanding~\cite{wang2013interpretable, rao2016multi}.
Furthermore, recent works~\cite{yang2022emotion, yang2024towards} introduce contextual understanding for visual emotion analysis, since emotions are conveyed through scene interplay, social interactions, and the overall environment.
Recent large language models (LLMs) have led to notable advancements in understanding text descriptions and generating corresponding responses~\cite{llama2, gpt4v}. With Chain-of-Thought (CoT) prompting~\cite{kojima2022large,li2024enhancing}, LLMs are improved towards deeper emotional understanding.
Inspired by these pioneering works, AIF models ensemble low-level and mid-level features as cues, leveraging emotional reasoning to accurately analyze nuanced emotions within text descriptions. High-level features are omitted since an image filter should preserve the fundamental content or narrative of images.

\subsection{Image Style Transfer.}
Early style transfer models focus on replicating individual styles through optimization-based methods~\cite{gatys2016image} or designing end-to-end convolutional neural networks~\cite{johnson2016perceptual, li2016precomputed}. 
Later, arbitrary style transfer models~\cite{huang2017arbitrary, li2017universal, chen2017stylebank, svoboda2020two} adaptively synthesize stylized results from any given reference image.
To improve the ability of contextual understanding, attention mechanisms are integrated~\cite{deng2022stytr2, liu2021adaattn, park2019arbitrary, wu2021styleformer}, establishing long-range dependencies between image patches.
In recent years, diffusion models~\cite{stablediffusion,controlnet} break new ground in producing high-fidelity images, which facilitates style transfer research to incorporate the generative capabilities of these models~\cite{chen2023artfusion, li2023stylediffusion, zhang2023inversion}. 
Meanwhile, text descriptions are gradually replacing traditional reference images, offering a more flexible and user-friendly approach to style transfer~\cite{clipstyler, yang2023zero, clva}. 
Compared to style transfer methods that rely on explicit color and texture features, an AIF model should reflect implicit emotional cues.

\section{Common setups for AIF Task} \label{sec:setup}
The AIF task enables users to reflect their thoughts and feelings, which requires the model to understand visually-abstract emotions from user-provided text descriptions and reflect them into visually-concrete images with appropriate colors and textures. 
We introduce the AIF dataset to provide comprehensive emotional data. 
Additionally, we formulate the AIF model with four key components.

\subsection{AIF Dataset Collection} \label{sec:dataset}
Although previous datasets (\eg, ArtEmis~\cite{achlioptas2021artemis} and ArtEmis v2~\cite{mohamed2022artemisv2}) connect visual elements, text descriptions, and emotions, their descriptions may emphasize the visual content over the evoked emotions of artistic images. This limitation motivates us to create the AIF dataset to prepare qualified data for training and evaluating relevant models. 
Specifically, we first merge all images from ArtEmis~\cite{achlioptas2021artemis} and ArtEmis v2~\cite{mohamed2022artemisv2}.
Then, we manually filter the unqualified text descriptions, remaining those that illustrate evoked emotions and discarding those that focus on visual details. 
After that, we ensure each image in the AIF dataset has an average of four to five such emotional descriptions, and categorize these descriptions according to Mikel's wheel~\cite{mikels2005emotional}. Due to the inherent subjectivity and ambiguity of emotions, we represent the evoked emotions of each image as a distribution across these categories, enabling the model to capture the relationships between emotions.

Finally, the AIF dataset consists of abundant aesthetic samples as anchor images that have diverse colors and textures, used as fixed points of visually-concrete references for visualizing colors and textures from visually-abstract emotions. Each image is paired with several corresponding text descriptions associated with the emotional category labels in Mikel's wheel~\cite{mikels2005emotional}.
We directly adopt the original emotional categories provided by the source datasets. To ensure the quality and reliability of these annotations, we conduct a second round of verification on a 10\% subset of our dataset.
After measuring the inter-rater agreement using Cohen’s Kappa~\cite{cohenkappa}, the result is 0.81, indicating almost perfect agreement and confirming the reliability of the annotations.
We split the dataset into a training set (69.6K images and 292.9K text descriptions) and an evaluation set (7.7K images and 32.5K text descriptions). To evaluate the performance of relevant methods, we use samples from the COCO-Stuff~\cite{coco-stuff} dataset as content images.

\subsection{AIF Model Formulation} \label{sec:formulation}
We systematically formulate an AIF model with the following four key components: 

\subsubsection{Inputs} \label{sec:formulation-0}
Content images $I^{\mathrm{cnt}} \in \mathbb{R}^{H \times W \times 3}$ to provide visual content and text descriptions $T^{\mathrm{des}} \in {\mathbb{N}^M}$ to evoke specific emotions, where $H$ and $W$ are the image size, and $M$ is the number of words in the text descriptions.

\subsubsection{Crafting architecture with content preservation}  \label{sec:formulation-1}
Content images $I^{\mathrm{cnt}}$ and text descriptions $T^{\mathrm{des}}$ are encoded by an image encoder $\mathcal{E}$ and a text encoder $\mathcal{T}$, respectively:
\begin{equation}
    Z^{\mathrm{img}} = \mathcal{E}(I^{\mathrm{cnt}}),   \qquad Z^{\mathrm{tex}} = \mathcal{T}(T^{\mathrm{des}}),
    \label{eq:formulation_image}
\end{equation}
where $Z^{\mathrm{img}} \in \mathbb{R}^{h \times w \times C_{\mathrm{img}}}$ and $Z^{\mathrm{tex}}  \in \mathbb{R}^{M \times C_{\mathrm{tex}}}$ are separately image tokens and text tokens. $h = H/f$, $w = W/f$, and $f$ is a hyper-parameter of the downsampling rate. $C_{\mathrm{img}}$ and $C_{\mathrm{tex}}$ are the numbers of embedding channels of image and text tokens, respectively. 
To capture long-range dependencies between tokens across different modalities, both the image tokens $Z^{\mathrm{img}}$ and text tokens $Z^{\mathrm{tex}}$ are fed into a generation model $G$ to preserve the overall structure of content images:
\begin{equation}
    Z^{\prime} = G(Z^{\mathrm{img}}, Z^{\mathrm{tex}}).
    \label{eq:formulation_generation}
\end{equation}

An image decoder $\mathcal{D}$ is further used to synthesize images $I^{\mathrm{out}} \in \mathbb{R}^{H \times W \times 3}$ that effectively reflect specific emotional responses from human observers:
\begin{equation}
    I^{\mathrm{out}} = \mathcal{D}(Z^{\prime}).
    \label{eq:formulation_decoder}
\end{equation}

\subsubsection{Learning high-dimensional emotional cues} \label{sec:formulation-2}
To understand inherent emotional properties, we augment the emotional expression of text tokens $Z^{\mathrm{emo}}$ by incorporating external emotional cues:
\begin{equation}
    Z^{\mathrm{emo}} = P(Z^{\mathrm{tex}}),
    \label{eq:formulation_prior}
\end{equation}
where $P$ represents an emotional augmentation module. We further replace $Z^{\mathrm{tex}}$ with $Z^{\mathrm{emo}}$ in~\cref{eq:formulation_generation} to leverage these emotional cues.
Since emotions are ambiguous and subjective, an emotional distribution loss $\mathcal{L}_{\mathrm{ed}}$ is required to capture the synthesized emotional cues, which ensures the evoked emotional distributions align with users' intentions.

\subsubsection{Creating concrete emotional visualization} \label{sec:formulation-3}
The AIF model is equipped with four losses to synthesize visually appealing images.
To ensure the synthesized images could evoke specific emotional responses, a sentiment metric loss $\mathcal{L}_{\mathrm{sm}}$ is required to learn relationships between emotions.
To create appropriate colors and textures, an anchor-based sentiment loss $\mathcal{L}_{\mathrm{as}}$ is required to learn artistic representations from corresponding anchor images.
Additionally, to improve the aesthetic quality of synthesized images, the aesthetic loss $\mathcal{L}_{\mathrm{ae}}$ is required to balance artistic style with content consistency.

\begin{figure*}[t]
  \centering
  \includegraphics[width=\linewidth]{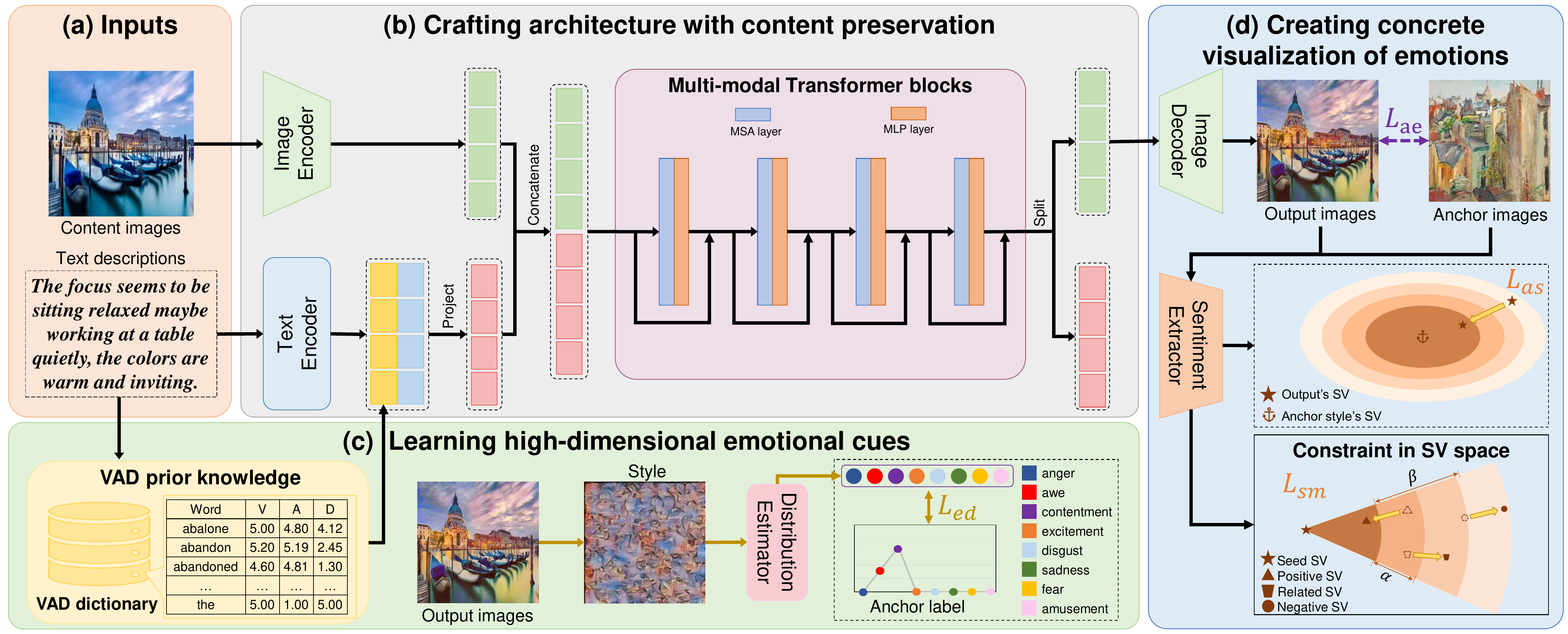}
  \caption{The pipeline of AIF-B~\cite{aif}. 
  (a)~Users input content images to provide visual content and text descriptions to evoke specific emotions. 
  (b)~Content images and text descriptions are encoded by the image encoder and text encoder, respectively. The encoded image and text tokens are fed into multi-modal transformer blocks, each comprising a Multi-headed Self-Attention (MSA) layer, a Multi-Layer Perceptron (MLP) layer, and residual connections (\cref{sec:b-architecture}). (c)~To understand inherent emotional properties, AIF-B leverages the VAD dictionary as emotional prior knowledge. The emotional distribution loss is introduced to capture high-dimensional emotional cues (\cref{sec:b-understanding}). 
  (d)~The sentiment metric loss and anchor-based sentiment loss are proposed to effectively reflect specific emotions (\cref{sec:b-emotion}). The aesthetic loss is further adopted to ensure the aesthetic quality and enrich the artistic style of synthesized images (\cref{sec:b-visualization}).} 
  \label{fig:AIFormer}
\end{figure*}

\section{Basic AIF model} \label{sec:aif-b}
In this section, AIF-B is introduced as an initial attempt based on the AIF model formulation (\cref{sec:formulation}).
We first present the multi-model transformer architecture (\cref{sec:b-architecture}). 
Then, we fetch the emotional cues from the VAD dictionary and design the emotional distribution loss (\cref{sec:b-understanding}).
Next, we introduce the sentiment loss and the anchor-based sentiment loss (\cref{sec:b-emotion}), along with the aesthetic loss (\cref{sec:b-visualization}), for effective emotional visualization.
Finally, we provide the details of training settings (\cref{sec:b-details}).

\subsection{Multi-modal Transformer} \label{sec:b-architecture}
As illustrated in~\cref{fig:AIFormer}~(b), given sampled batches, the multi-modal transformer separately encodes text descriptions and content images, and captures their dependencies in the subsequent transformer blocks. Specifically, the multi-modal transformer can be divided into four components.

\noindent \textbf{Image encoder.} Content images $I^{\mathrm{cnt}}$ are split into image patches and a transformer-based image encoder is adopted to extract image tokens $Z^\mathrm{img} = [Z^\mathrm{img}_1, \dots, Z^\mathrm{img}_N] \in \mathbb{R}^{N \times C_\mathrm{img}}$, where $N$ is the number of image patches and $C_{\mathrm{img}}$ is the number of embedding channels.

\noindent \textbf{Text encoder.} Text descriptions are encoded with pre-trained BERT~\cite{bert} 
to form text tokens $Z^{\mathrm{tex}} = [Z^{\mathrm{tex}}_{1}, \dots, Z^{\mathrm{tex}}_{M}] \in \mathbb{R}^{M \times C_{\mathrm{tex}}}$, where $M$ is the number of words and $C_{\mathrm{tex}}$ is the number of embedding channels.

\noindent \textbf{Multi-modal transformer.} A stack of MLP layers further projects the image and text tokens into a shared latent space.
Then, modal-type embeddings $Z^\mathrm{typ}_\mathrm{0}$ and $Z^\mathrm{typ}_\mathrm{1}$ are introduced to distinguish token modalities:
\begin{equation}
    \hat{Z}^\mathrm{img} = \mathrm{MLP}(Z^\mathrm{img})+Z^\mathrm{typ}_\mathrm{0}, \ 
    \hat{Z}^\mathrm{tex} = \mathrm{MLP}(Z^\mathrm{tex})+Z^\mathrm{typ}_\mathrm{1}, 
    \label{eq:trans_1}
\end{equation}
These embeddings are then concatenated as the initial input of the transformer:
\begin{equation}
    Z_{0} = [\hat{Z}^{\mathrm{img}}_{1}, \dots, \hat{Z}^{\mathrm{img}}_{N}, \hat{Z}^{\mathrm{tex}}_{1}, \dots, \hat{Z}^{\mathrm{tex}}_{M}] \in \mathbb{R}^{(N+M) \times C_{0}},
    \label{eq:trans_2}
\end{equation}
where $C_{0}$ is the number of channel.
The adopted transformer consists of $L$ transformer blocks~\cite{vit}, and each block includes a self-attention layer, an MLP layer, and two residual connections. The computation within each block is formulated as:
\begin{align}
    [\bar{Z}_i] &= \mathrm{MSA}(\mathrm{LN}([Z_{i-1}])) + [Z_{i-1}], &\  i \in \{ 1,\dots,L \}  \\
    [Z_i] &= \mathrm{MLP}(\mathrm{LN}([\bar{Z}_i])) + [\bar{Z}_i], &\  i \in \{ 1,\dots,L \}
    \label{eq:trans_4}
\end{align}
where LN means the LayerNorm.

\noindent \textbf{Image decoder.} A three-layer CNN is used as the image decoder to synthesize results $I^{\mathrm{out}}$, which alleviates the grid artifacts caused by the patch partition of images.

\subsection{External Emotional Cues} \label{sec:b-understanding}
As illustrated in~\cref{fig:AIFormer}~(c), the VAD dictionary, as an emotional augmentation module, augments emotional expression. The emotional distribution loss is employed to evoke emotions accurately.

\noindent \textbf{VAD prior knowledge.} Affective tokens are fetched from the VAD dictionary~\cite{mohammad2018vad} and used as the emotional cues to extract the inherent emotional properties of each word (\ie, valence, arousal, and dominance). Specifically, valence indicates the positivity or negativity of the emotion, arousal reflects the intensity or excitement level of the emotion, and dominance represents the degree of control or influence one feels with the emotion. Words not included in the VAD dictionary are assigned neutral values to ensure the completeness and consistency of emotional understanding.
These tokens are denoted as $Z^{\mathrm{vad}} \in \mathbb{R}^{M \times C_\mathrm{vad}}$, where $C_\mathrm{vad}$ is the number of embedding channels.
To augment the emotional expression, text tokens are concatenated with affective tokens as $Z^{\mathrm{emo}} = [Z^{\mathrm{tex}}; Z^{\mathrm{vad}}] \in \mathbb{R}^{M \times (C_\mathrm{tex}+C_\mathrm{vad})}$, replacing $Z^{\mathrm{tex}}$ in~\cref{eq:trans_1} and~\cref{eq:trans_2}.

\noindent \textbf{Emotional distribution loss.} To effectively capture synthesized emotional cues, an emotional distribution loss is proposed. Since emotions are inherently ambiguous and subjective, a single image can evoke a diverse range of emotional responses even within the same individual. 
Instead of categorizing a single dominant emotion for each image, the emotional distribution loss aims to estimate the comprehensive emotional distribution. 
Specifically, a VGG-based distribution estimator $\varphi$ is used to extract color and style features through the mean and variance of the feature representations. The emotional distribution loss is defined using the Kullback-Leibler (KL) divergence~\cite{kullback1951information}, which minimizes the discrepancy between the emotional distributions of synthesized images and the ground truth:
\begin{equation}
    \mathcal{L}_{\mathrm{ed}} = \sum_{i=1}^{N_{\mathrm{cat}}} d_{i} \mathrm{ln} \frac{d_{i}}{\varphi(I^{\mathrm{out}})_{i}},
    \label{eq:aiformer_ed}
\end{equation}
where $N_{\mathrm{cat}}$ is the number of emotional categories, $\varphi(I^{\mathrm{out}})_{i}$ and $d_{i}$ are the value of $i$-th category of the estimated distribution and the ground truth, respectively.

\subsection{Emotional Reflection Loss} \label{sec:b-emotion}
As demonstrated in~\cref{fig:AIFormer}~(d), the sentiment metric loss $\mathcal{L_{\mathrm{sm}}}$ and anchor-based sentiment loss $\mathcal{L_{\mathrm{as}}}$ are proposed to enable synthesized images to effectively reflect specific emotions.

\noindent \textbf{Sentiment metric loss.}
AIF-B learns relationships between emotions in three steps:
\textit{(i)} Following Yang~\cite{yang2018retrieving}, a sentiment extractor is built. A VGG network is used to extract multi-level features of synthesized images. These features are then projected through a convolutional layer and used to calculate multiple Gram matrices at multiple feature levels. 
The sentiment vector $V$ is then defined by concatenating selected elements from these Gram matrices:
\begin{equation}
    V = \mathrm{Concat}_{i \in \left\{1, \dots, N_{\mathrm{gram}} \right\} \cup j \in \left\{1, \dots, N_{\mathrm{lev}} \right\}} (\Phi_{i, j}),
    \label{eq:sentivector}
\end{equation}
where $\Phi_{i,j}$ means the $i$-th upper triangular elements in the Gram matrix at $j$-th feature level, $N_{\mathrm{gram}}$ is the number of elements, and $N_{\mathrm{lev}}$ is the number of levels.
\textit{(ii)} Based on Mikel's wheel~\cite{mikels2005emotional} that defines relationship between emotions, and given randomly sampled text descriptions $T^\mathrm{sed}$, tuples of text samples $[T^{\mathrm{sed}}, T^{\mathrm{pos}}, T^{\mathrm{rel}}, T^{\mathrm{neg}}]$ are constructed. Here, $T^{\mathrm{pos}}$ represents positive samples from the same emotional region, $T^{\mathrm{rel}}$ represents related samples from adjacent emotional regions, and $T^{\mathrm{neg}}$ represents negative samples from the opposite emotional region. The corresponding sentiment vectors are denoted as $[V^{\mathrm{sed}}, V^{\mathrm{pos}}, V^{\mathrm{rel}}, V^{\mathrm{neg}}]$.
\textit{(iii)} Defining the emotional distance $\mathrm{F}_{\mathrm{dis}}$ as the minimum number of steps between emotion regions in the Mikel's wheel~\cite{mikels2005emotional}, the distance between sentiment vectors is formulated as: 
$\mathrm{F}_{\mathrm{sw}}(V_{i}, V_{j}) = \frac{\|V_{\mathrm{i}} - V_{\mathrm{j}}\|^2}{\mathrm{F}_\mathrm{dis}(V_{i}, V_{j})}$.
As a result, the sentiment metric loss is formulated in a metric-learning manner:
\begin{align}
     \!\!\!\! \mathcal{L}_{\mathrm{sm}} \!
     \!= & \mathrm{max}(\mathrm{F}_\mathrm{sw}(V^{\mathrm{sed}}, V^{\mathrm{pos}}) \!-\! \mathrm{F}_{\mathrm{sw}}(V^{\mathrm{sed}}, V^{\mathrm{rel}}) \!+\! \alpha), 0) \nonumber \\
    + & \mathrm{max}(\mathrm{F}_{\mathrm{sw}}(V^{\mathrm{sed}}, V^{\mathrm{rel}}) \!-\! \mathrm{F}_{\mathrm{sw}}(V^{\mathrm{sed}}, V^{\mathrm{neg}}) \!+\! \beta), 0), \!\!
    \label{eq:sw}
\end{align}
where $\alpha=0.02$ and $\beta=0.01$ are hyper-parameters that control margins between sentiment vectors.

\noindent \textbf{Anchor-based sentiment loss.} 
To ensure that synthesized images have appropriate colors and textures, the sentiment extractor is further used to obtain sentiment vectors for each synthesized images and their corresponding anchor images (denoted as $V^{\mathrm{out}}$ and $V^{\mathrm{acr}}$). 
By minimizing the mean squared error between these vectors, the model aligns the synthesized images with the desired artistic representations:
\begin{equation}
\mathcal{L}_{\mathrm{as}} = \| V^{\mathrm{out}} - V^{\mathrm{acr}} \|_{2}.
\label{eq:abs}
\end{equation}

\subsection{Visual Aesthetics Loss} \label{sec:b-visualization}
As depicted in~\cref{fig:AIFormer}~(d), the aesthetic loss $\mathcal{L}_{\mathrm{ae}}$ is designed to improve the aesthetic quality of synthesized images. In practice, AIF-B defines aesthetic loss as a composite function, including content loss $\mathcal{L}_{\mathrm{c}}$~\cite{huang2017arbitrary}, style loss $\mathcal{L}_{\mathrm{s}}$~\cite{deng2022stytr2}, GAN loss $\mathcal{L}_{\mathrm{GAN}}$~\cite{stackgan++}, and identity loss $\mathcal{L}_{\mathrm{id}}$~\cite{park2019arbitrary}, detailed as follows:

\noindent \textbf{Content loss.} To preserve the visual features and overall structure, the content loss uses a pre-defined VGG network~\cite{vgg} to extract multi-level features and narrows the squared error between corresponding features at each level as:
\begin{align}
    \mathcal{L}_{\mathrm{c}} = \sum_{i} \| \phi^{\mathrm{out}}_{i} - \phi^{\mathrm{cnt}}_{i} \|_2,
    \label{eq:content}
\end{align}
where $\phi^{\mathrm{out}}_{i}$ and $\phi^{\mathrm{cnt}}_{i}$ are extracted features of synthesized images and content images at $i$-th level, respectively.

\noindent \textbf{Style loss.} 
Following the style transfer methods \cite{deng2022stytr2, huang2017arbitrary}, we represent the image style using the mean and variance of extracted feature maps. To transfer the artistic style of images, the style loss minimizes the style difference between feature statistics of synthesized images and anchor images at each level as:
\begin{align}
    \!\!\!\! \mathcal{L}_{\mathrm{s}} = \sum_{i} \|\mu(\phi^{\mathrm{out}}_{i}) - \mu(\phi^{\mathrm{acr}}_{i})\|_2  + \|\sigma(\phi^{\mathrm{out}}_{i}) - \sigma(\phi^{\mathrm{acr}}_{i})\|_2,\!\!\!\!
    \label{eq:style}
\end{align}
where $\mu$ and $\sigma$ denote the mean and variance functions, respectively. $\phi^{\mathrm{acr}}_{i}$ is extracted features of anchor images.

\noindent \textbf{GAN loss.} 
To ensure consistency between synthesized images and text descriptions while maintaining aesthetic quality, a 
conditional-unconditional discriminator is introduced:
\begin{align}
    \label{eq:GAN}
    \!\! \mathcal{L}_{\mathrm{GAN}} =  \log D(I^{\mathrm{acr}}) & + \log \big (1-D(G(I^{\mathrm{cnt}}, Z^{\mathrm{emo}})) \big) \\
    + \log D(I^{\mathrm{acr}},Z^{\mathrm{emo}})  & + \log \big(1-D (G(I^{\mathrm{cnt}}, Z^{\mathrm{emo}}), Z^{\mathrm{emo}}) \big), \!\! \nonumber
\end{align}
where $D$ and $G$ are the discriminator and generator.

\noindent \textbf{Identity loss.} 
To learn richer semantic representations (\eg, color, texture, and visual content), anchor images and corresponding text descriptions are fed into the AIF model to synthesize identity images. The identity loss ensures consistency between identity images and their original anchor images:
\begin{align}
    \mathcal{L}_{\mathrm{id}} = \| I^{\mathrm{idt}} - I^{\mathrm{acr}} \|_{2} +  \lambda \sum_{i} \| \phi^\mathrm{idt}_{i} - \phi^\mathrm{acr}_{i} \|_2,
    \label{eq:id}
\end{align}
where $I^{\mathrm{idt}}$ and $\phi^{\mathrm{idt}}_i$ are identity images and extracted features at $i$-th level, respectively. $I^{\mathrm{acr}}$ means corresponding anchor images. $\lambda=0.01$ is a hyper-parameter.

Consequently, the aesthetic loss in AIF-B is defined as:
\begin{align}
    \mathcal{L}_{\mathrm{ae}} = \lambda_{\mathrm{c}} \mathcal{L}_{\mathrm{c}} + \lambda_{\mathrm{s}} \mathcal{L}_{\mathrm{s}} + \lambda_{\mathrm{gan}} \mathcal{L}_{\mathrm{GAN}} + \lambda_{\mathrm{id}} \mathcal{L}_{\mathrm{id}},
    \label{eq:ae}
\end{align}
where $\lambda_{\mathrm{c}} = 5$, $\lambda_{\mathrm{s}} = 0.3$, $\lambda_{\mathrm{gan}} = 3$, and $\lambda_{\mathrm{id}} = 2$ denote the 
weights for losses.

\begin{figure*}[t]
  \centering
  \includegraphics[width=\linewidth]{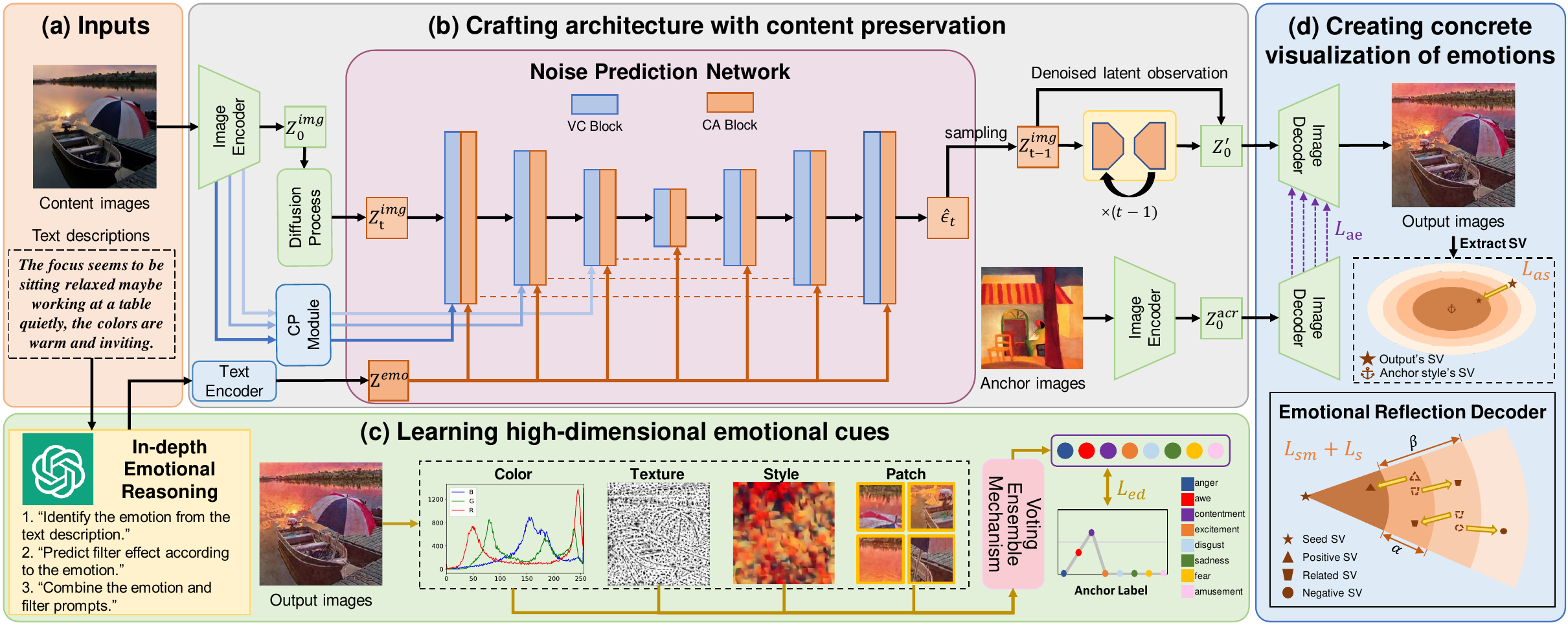}
  \caption{The pipeline of AIF-D. 
  (a)~Users input content images to provide visual content and text descriptions to evoke specific emotions.
  (b)~Content images and text descriptions are fed into the image encoder and text encoder to extract image tokens and text tokens, respectively.
  The noise prediction network estimates the noise at each diffusion step. Within the downsampling modules, each Vanilla Convolution (VC) block before the Cross-Attention (CA) blocks is equipped with a Content Preservation (CP) module. These CP modules integrate the context of content images to preserve high-frequency details (\cref{sec:d-architecture}).
  (c)~An LLM and CoT prompting are used to handle complex emotional expressions with in-depth emotional reasoning. The voting ensemble mechanism is introduced to evaluate emotional distributions from different points of view and ensembles low-level (colors and texture) and mid-level features (image style and patch features) to enable accurate analysis of evoked emotions (\cref{sec:d-understanding}). 
  (d)~To enable accurate visualization of emotions, we propose an emotional reflection strategy to enhance the emotional understanding of the image decoder, using the sentiment metric loss and anchor-based sentiment loss (\cref{sec:d-emotion}). Finally, images are synthesized using a redesigned aesthetic loss, achieving a balance between artistic style and content consistency (\cref{sec:d-visualization}).}
  \label{fig:pipeline}
\end{figure*}

\vspace{-1mm}
\subsection{Learning and training details} \label{sec:b-details}
\noindent\textbf{Losses.} AIF-B is trained by solving a minimax optimization problem involving full objective losses:
\begin{align}
    \max_{D} \min_{G}  
    \lambda_{\mathrm{ed}} \mathcal{L}_{\mathrm{ed}} +
    \lambda_{\mathrm{sm}} \mathcal{L}_{\mathrm{sm}} +
    \lambda_{\mathrm{as}} \mathcal{L}_{\mathrm{as}} +
    \lambda_{\mathrm{ae}} \mathcal{L}_{\mathrm{ae}}, \nonumber
\end{align}
where $\lambda_{\mathrm{ed}}=140$, $\lambda_{\mathrm{sm}}=30$, $\lambda_{\mathrm{as}}=600$, and $\lambda_{\mathrm{ae}}=1$ are hyper-parameters, which are determined empirically.

\noindent\textbf{Training Details.} AIF-B is trained on $4$ NVIDIA TITAN RTX GPUs for $80$K iterations with batch size $24$ for $30$ hours. We use the Adam optimizer minimizes losses with the warm-up adjustment strategy~\cite{warm-up}.

\section{Deeper AIF model} \label{sec:aif-d}
Although AIF-B takes the first step to concrete visualization of emotions, it still suffers from blurry details, keyword blindness, and overly-stylized effects. In this section, we present AIF-D as an extension of AIF-B, advancing towards deeper emotional reflection. 
For clarity, we follow the same presentation structure as AIF-B (\cref{sec:aif-b}) to describe AIF-D. 
We begin by introducing the diffusion-model backbone with a content preservation module (\cref{sec:d-architecture}).
Then, we leverage an LLM for in-depth emotional reasoning while ensembling emotional cues from different perspectives (\cref{sec:d-understanding}).
We improve the emotional understanding of the image decoder based on the sentiment metric and anchor-based sentiment loss (\cref{sec:d-emotion}).
The aesthetic loss is further proposed to balance artistic style and content consistency (\cref{sec:d-visualization})
Finally, we provide the details of training settings (\cref{sec:d-details}).

\subsection{Diffusion-model backbone} \label{sec:d-architecture}
As shown in~\cref{fig:pipeline}~(b), AIF-D is built upon the pre-trained latent diffusion model~\cite{stablediffusion} to leverage its rich generative priors, rather than training a multi-modal transformer from scratch (\cref{sec:b-architecture}). The model can be divided into four components.

\noindent \textbf{Condition encoders.}
A pre-trained Variational Autoencoder (VAE)~\cite{vae} is adopted as the image encoder $\mathcal{E}$, converting content images $I^{\mathrm{cnt}}$ into image tokens $Z^{\mathrm{img}}$.
Text descriptions $T^{\mathrm{des}}$ are processed through the CLIP~\cite{clip}, as the text encoder $\mathcal{T}$, to generate text tokens $Z^{\mathrm{tex}}$.
These tokens are then injected into the latent space of a UNet-based noise prediction network~$\epsilon_\theta$, which incorporates the cross-attention mechanism~\cite{attn}.

\noindent \textbf{Content preservation module.}
As the condition encoder compresses the image, high-frequency details are mostly lost, resulting in blurry details and reduced content consistency. To address this degradation, the integration of a content preservation module is proposed within the noise prediction network, leveraging intermediate encoder features as the additional context of the content image. 
Specifically, this module injects multi-scale features from the image encoder into a modified U-Net architecture~\cite{stablediffusion}, serving as contextual anchors that guide the noise prediction. By injecting these intermediate features into corresponding stages of the U-Net backbone, the content preservation module preserves high-frequency details and improves content consistency. This process could be expressed as:
\begin{equation}
    f_{i}^\prime = f_{i} + \mathrm{F}_{\mathrm{cp}}(\mathcal{E}_{i}), \qquad  i \in \{1, \dots, N_{\mathrm{enc}} \},
\end{equation}
where $f_{i}$ and $\mathcal{E}_{i}$ are the features in the U-Net backbone and the image encoder of the $i$-block respectively, $\mathrm{F}_{\mathrm{cp}}$ means a stack of downsampling convolution, and $N_{\mathrm{enc}}$ is the number of blocks in the image encoder.

\noindent \textbf{Forward process.} Given the property of the forward process in diffusion models~\cite{ddpm, ddim}, the forward process plays a critical role in transforming the image tokens into a latent space, learning a robust generative pathway. 
Specifically, let $Z^{\mathrm{img}}_{0}$ denote the encoded image features $Z^{\mathrm{img}}$ at timestep $t=0$. The noised sample $Z^{\mathrm{img}}_{t}$ can be expressed as a linear combination of $Z^{\mathrm{img}}_{0}$ and a noise variable $\epsilon$ as the diffusion process:
\begin{equation}
    Z^{\mathrm{img}}_{t} = \sqrt{\alpha_{t}} Z^{\mathrm{img}}_{0} + \sqrt{1-\alpha_{t}} \epsilon, \label{eq:forward}
\end{equation}
where $\epsilon \sim \mathcal{N}(0, 1)$ is the Gaussian noise and $\alpha_{t}$ is the parameter for controlling noise levels.

\noindent \textbf{Backward process.}
The noise prediction network aims to iteratively refine the noised sample $Z^{\mathrm{img}}_{t}$ until it converges to the clear content latent code $Z_{0}^{\prime}$:
\begin{small}
\begin{equation}
    Z^{\mathrm{img}}_{t-1} = \frac{1}{\sqrt{\alpha_{t}}} \big (Z^{\mathrm{img}}_{t} - \frac{1-\alpha_{t}}{\sqrt{1-\bar{\alpha_{t}}}}\epsilon_\theta(Z^{\mathrm{img}}_{t}, t, Z^{\mathrm{tex}}, Z^{\mathrm{img}}_{0}) \big ) \! + \sigma_{t}\epsilon,  \label{eq:back}
\end{equation}
\end{small}
\vspace{-3mm}\\
where $\bar{\alpha}_{t} = \prod_{s=1}^{t}\alpha_{s}$ controls the amount of noise, $Z^{\mathrm{tex}}$ denotes text tokens, $\sigma_{t}$ defines the standard deviation of the noise. Finally, a pre-trained VAE image decoder $\mathcal{D}$ is used to map the latent code back into the pixel space and present the synthesized image as $I^{\mathrm{out}} = \mathcal{D}(Z_{0}^{\prime})$.

\begin{figure*}[t]
  \centering
  \includegraphics[width=\linewidth]{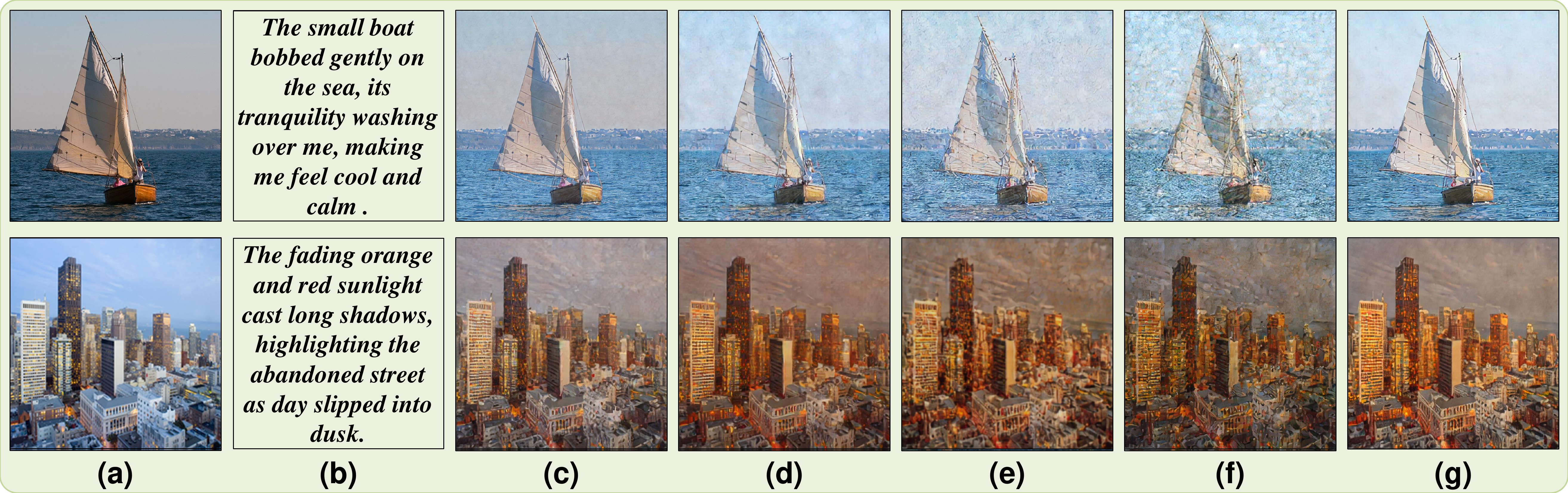}
  \caption{Visualization of applying the texture mapping loss across different blocks of the image decoder.
  (a) User-provided content images. (b) Text descriptions that reflect thoughts and feelings. (c)-(f) Results from applying the texture mapping loss at progressively later decoder blocks. (g) Results of AIF-D with appropriate artistic style and content consistency by combining losses across blocks.}
  \label{fig:texturemapping}
\end{figure*}

\subsection{Ensembling Emotional Cues} \label{sec:d-understanding}
As illustrated in~\cref{fig:pipeline}~(c), AIF-D replaces the VAD dictionary (\cref{sec:b-understanding}) with an LLM and CoT prompting as the emotional augmentation module to facilitate in-depth emotional reasoning.
Instead of estimating emotional distribution solely from style (\cref{sec:b-understanding}), the voting ensemble mechanism extracts image features from
different perspectives to ensure the synthesized results evoke the corresponding emotions.

\noindent \textbf{In-depth emotional reasoning.} 
Relying solely on individual emotional words can lead to keyword blindness and failure to capture complex emotional expressions.
Towards deeper understanding of emotions, an LLM~\cite{gpt4v} and the CoT prompting~\cite{kojima2022large,li2024enhancing} are introduced to facilitate in-depth emotional reasoning. 
Specifically, the CoT prompting process enhances user-provided text descriptions into a rich emotional prompt through three sequential steps:
    \textit{(i)} First, we prompt the LLM~\cite{gpt4v} to analyze the text description and extract emotional keywords (\eg, calmness, contentment, and enjoyment).
    \textit{(ii)} Next, using these identified keywords, we guide the LLM~\cite{gpt4v} to formulate an enhancement directive for the image content (\eg, the room's contentment and enjoyment).
    \textit{(iii)} Finally, the LLM combines the original text description with the enhancement directive to generate the rich emotional prompt for the subsequent process.
    We present the details of the reasoning process in the Supp.

\noindent \textbf{Voting ensemble mechanism.} 
People may experience a variety of emotions when looking at the same image because they have different perceptual preferences, cognitive styles, and aesthetic tendencies. Consequently, rather than solely relying on VGG-based emotional distribution estimation, a voting ensemble mechanism is designed to simulate people viewing images from different perspectives, aggregating votes to accurately analyze the evoked emotions. Specifically, the emotional distribution is estimated separately using low-level features~\cite{zhao2014exploring, machajdik2010affective} and mid-level features~\cite{yang2018weakly, park2020swapping}:
\textit{(i)} \textbf{Color.} Color histograms are independently computed for each of the RGB color channels in images.
\textit{(ii)} \textbf{Texture.} The gray-level co-occurrence matrix (GLCM)~\cite{zhao2014exploring} is computed, followed by the extraction of handcrafted features.
\textit{(iii)} \textbf{Style.} A distribution estimator $\varphi$ extracts style features (\cref{sec:b-understanding}).
\textit{(iv)} \textbf{Patch.} The image is randomly cropped~\cite{park2020swapping} and the patch features are extracted. 
Consequently, $\phi^{j}(I^{\mathrm{out}})_{i}$ is defined as the value of the $i$-th category within the distribution estimated by the $j$-th classifier:
\begin{equation}
\begin{split}
    \varphi(I^{\mathrm{out}})_{i} = \frac{\sum_{j} \phi^{j}(I^{\mathrm{out}})_{i} \times w^{j}}{\sum_{j} w^{j}},
    \label{eq:ed}
\end{split}
\end{equation}
where ensemble weights $w^{j}$ are the accuracies of classifier models from each perspective, evaluated on a hold-out validation set.
Then, the emotional distribution loss $\mathcal{L}_{\mathrm{ed}}$ (\cref{eq:aiformer_ed}) is adopted to learn high-dimensional emotional cues effectively.

\subsection{Emotional Reflection Strategy} \label{sec:d-emotion}
As demonstrated in~\cref{fig:pipeline}~(d), the sentiment metric loss and anchor-based sentiment loss are reused from AIF-B (\cref{sec:b-emotion}).
To effectively learn both emotional and artistic priors, sentiment metric loss $\mathcal{L_{\mathrm{sm}}}$ is combined with the style loss $\mathcal{L_{\mathrm{s}}}$ to improve the emotional reflection of the image decoder.
Additionally, anchor-based sentiment loss $\mathcal{L_{\mathrm{as}}}$ is maintained to ensure the synthesized results evoke specific emotional responses.

\noindent \textbf{Emotion reflection decoder.}
Observing that images synthesized by the pre-trained image decoder~\cite{stablediffusion} cannot effectively produce colors and textures that align with the intended emotions, AIF-D enhances the emotional understanding of the decoder to appropriately present visually compelling artistic features.
Specifically, sentiment vectors $[V^{\mathrm{sed}}, V^{\mathrm{pos}}, V^{\mathrm{rel}}, V^{\mathrm{neg}}]$ are extracted from the images corresponding to text descriptions $[T^{\mathrm{sed}}, T^{\mathrm{pos}}, T^{\mathrm{rel}}, T^{\mathrm{neg}}]$. Defining the emotional distance as $\mathrm{F}_{\mathrm{sw}}(V_{i}, V_{j}) = \frac{\|V_{\mathrm{i}} - V_{\mathrm{j}}\|^2}{\mathrm{F}_\mathrm{dis}(V_{i}, V_{j})}$, the sentiment loss $\mathcal{L}_{\mathrm{sm}}$ (\cref{eq:sw}) is calculated to capture the relationships between emotions. To create appropriate colors and textures that reflect intended emotions, style loss $\mathcal{L}_{\mathrm{s}}$ (\cref{eq:style}) is adopted to minimize the style difference between synthesized results and anchor images. 
As a result, the image decoder is fine-tuned by anchoring visually abstract emotions to visually concrete images through a combined loss:
\begin{equation}
    \mathcal{L}_{\mathrm{ft}} = \mathcal{L}_{\mathrm{sm}} + \lambda \mathcal{L}_{\mathrm{s}},
\end{equation}
where hyper-parameter is set as $\lambda = 0.01$. Note that this process is completed before the training of the noise prediction network. The parameters of the image decoder are then fixed to preserve the learned emotional and artistic priors.

\noindent \textbf{Anchor-based sentiment loss.} Additionally, the anchor-based sentiment loss (\cref{eq:abs}) is adopted to minimize the distance between sentiment vectors of the synthesized results and their corresponding anchor images. Within the diffusion-model backbone, this remains effective in ensuring the synthesized results evoke specific emotional responses from human observers.

\subsection{Visual Aesthetics Balance} \label{sec:d-visualization}
As shown in~\cref{fig:pipeline}~(d), the aesthetic loss $\mathcal{L}_{\mathrm{ae}}$ is redesigned for the diffusion-based backbone to address the overly-stylized effects of AIF-B (\cref{sec:b-visualization}). 
The redesigned loss emphasizes balancing artistic style and content consistency while visualizing emotions as appropriate colors and textures. AIF-D defines aesthetic loss as a combination of the diffusion model loss $\mathcal{L}_{\mathrm{dm}}$ and the texture mapping loss $\mathcal{L}_{\mathrm{tm}}$, detailed as follows:

\noindent \textbf{Diffusion model loss.} 
Guided by the text tokens carrying emotional cues $Z^{\mathrm{emo}}$ and context of image tokens $Z^{\mathrm{img}}$, the MSE loss is adopted to minimize the discrepancy between the learned image distribution and the target distribution for arbitrary noised latent code:

\vspace{-2mm}
\begin{small}
\begin{equation}
    \mathcal{L}_{\mathrm{dm}} \! = \! \mathbb{E}_{t,Z^{\mathrm{img}}_{0},\epsilon \sim \mathcal{N}(0,1)} \big[\| \epsilon_{t} \! - \! \epsilon_{\theta}(Z^{\mathrm{img}}_{t}, t, Z^{\mathrm{emo}}, Z^{\mathrm{img}}_0) \|^{2}\big].
    \label{eq:dm}
\end{equation}
\end{small}

\noindent \textbf{Texture mapping Loss.} 
The texture mapping loss is presented to leverage the learned priors and the extracted decoding features.
Denote the denoised latent observation as:
\begin{equation}
    Z^{\prime} = {\big(Z_{t}^{\mathrm{img}} - \sqrt{1-\alpha_{t}} \epsilon_{\theta}(Z_{t}^{\mathrm{img}}, t, Z^{\mathrm{emo}}, Z^{\mathrm{img}}_0)\big)}/{\sqrt{\alpha_{t}}}.
\end{equation}
\vspace{-4mm} \\
Anchor images are then fed into the image encoder and decoder to extract multi-scale decoding features as guidance: 
\begin{equation}
    \mathcal{L}_{\mathrm{tm}}^{i} =  \mathrm{F}_{\mathrm{dif}} \big (\mathcal{D}_{i}(Z^{\mathrm{acr}}), \mathcal{D}_{i}(Z^{\prime}) \big),
\end{equation}
where $\mathrm{F}_{\mathrm{dif}}(a, b) = \|\mu(a) - \mu(b)\|_2  + \|\sigma(a) - \sigma(b)\|_2$ measures the difference between decoding features, $\mathcal{D}_{i}$ means features extracted from the $i$-th decoding block, and $Z^{\mathrm{acr}} = \mathcal{E}({I^{\mathrm{acr}}})$ represents encoded anchor tokens. 
To demonstrate the influence of applying the texture mapping loss at different decoder blocks, the synthesized results are visualized in~\cref{fig:texturemapping}. 
As the loss is applied to progressively later blocks in the decoder, the artistic features become stronger, accompanied by increased content distortion. Therefore, the texture mapping loss is calculated across multiple blocks to balance artistic style and content consistency:
\begin{equation}
    \mathcal{L}_{\mathrm{tm}} = \sum_{i} \gamma^{i+1} \mathcal{L}^{i}_{\mathrm{tm}},
    \label{eq:tm}
\end{equation}
where $\gamma = 0.3$ is a hyper-parameter.

Consequently, the aesthetic loss in AIF-D is defined as:
\begin{align}
    \mathcal{L}_{\mathrm{ae}} = \lambda_{\mathrm{dm}} \mathcal{L}_{\mathrm{dm}} + \lambda_{\mathrm{tm}} \mathcal{L}_{\mathrm{tm}},
    \label{eq:d-ae}
\end{align}
where $\lambda_{\mathrm{dm}} = 1$ and $\lambda_{\mathrm{tm}} = 0.001$ denote the weights for losses.

\subsection{Learning and training details} \label{sec:d-details}
\noindent \textbf{Losses.} AIF-D is trained by combining the emotional distribution loss~$\mathcal{L}_{\mathrm{ed}}$, anchor-based sentiment loss $\mathcal{L}_{\mathrm{as}}$, and aesthetic loss~$\mathcal{L}_{\mathrm{ae}}$ into the full objective loss for training the noise prediction network:
\begin{align}
    \mathcal{L}_{\mathrm{total}} = 
     \lambda_{\mathrm{ed}} \mathcal{L}_{\mathrm{ed}} + \lambda_{\mathrm{as}} \mathcal{L}_{\mathrm{as}} + \lambda_{\mathrm{ae}} \mathcal{L}_{\mathrm{ae}}
\end{align}
where $\lambda_{\mathrm{ed}}=10$,  $\lambda_{\mathrm{as}}=10$, and $\lambda_{\mathrm{ae}} = 1$ are hyper-parameters, which are determined empirically.

\noindent \textbf{Training details.}
The image decoder in AIF-D is first fine-tuned for only 2 epochs with a batch size of 4, a strategy similar to Imagic~\cite{Imagic}.
Subsequently, the parameters of the image decoder are fixed, and the noise prediction network is trained for 4 epochs with a batch size of 2, requiring approximately 68 hours with 2 NVIDIA GeForce 3090 GPUs. 
To supervise the synthesizing process conditioned on text descriptions, the classifier-free guidance~\cite{classifier-free} is applied. 
Adam optimizer~\cite{adam} is employed with learning rates of $1 \times 10^{-6}$ for the image decoder and the noise prediction network.

\begin{figure*}[t]
  \centering
  \includegraphics[width=\linewidth]{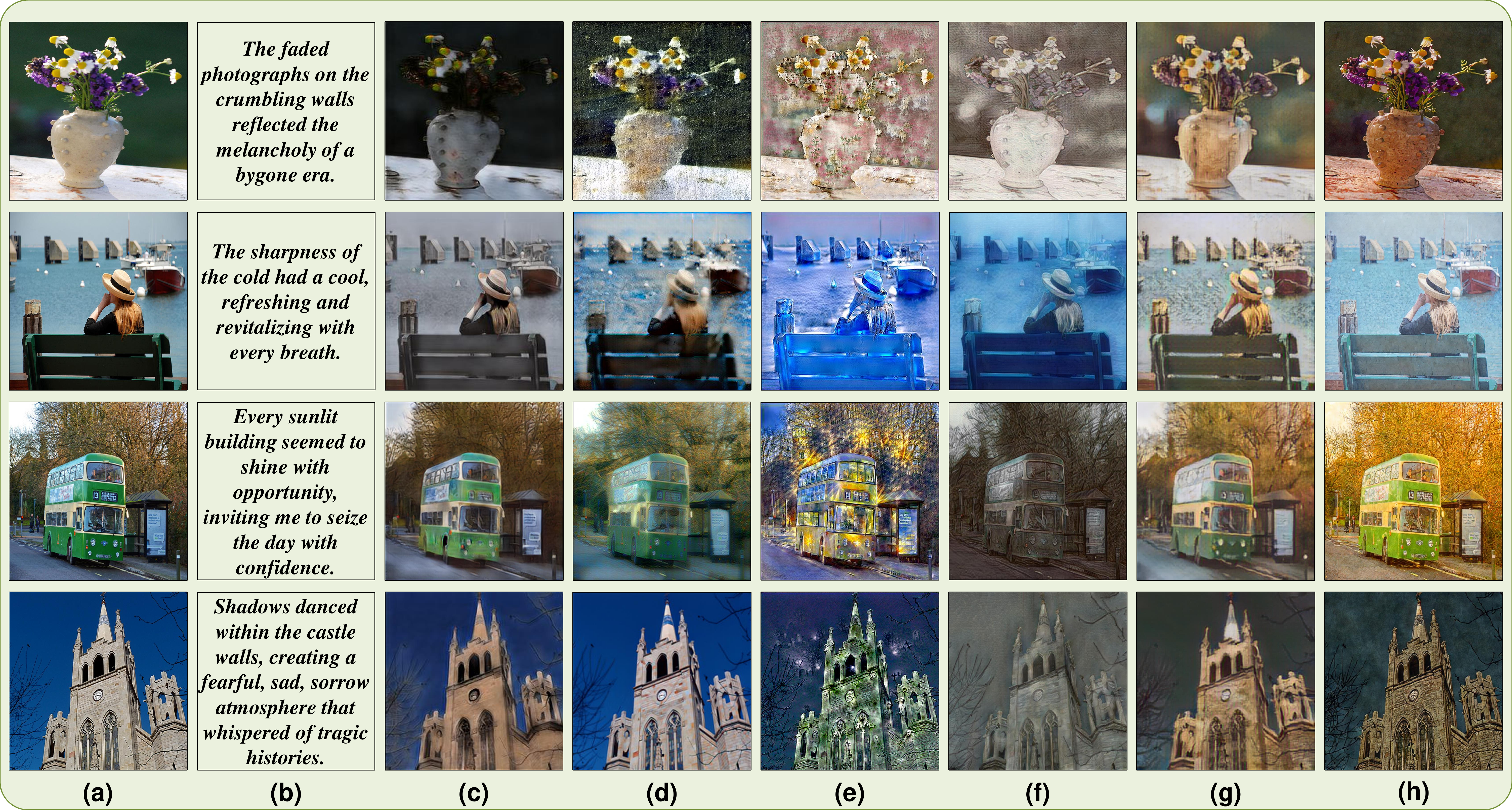}
  \caption{Qualitative comparison results with state-of-the-art methods. (a)~User-provided content images. (b)~Text descriptions that reflect thoughts and feelings. (c)~ManiGAN~\cite{manigan} (d)~DiffusioinCLIP~\cite{diffusionclip}. (e)~CLIPstyler~\cite{clipstyler}. (f)~CLVA~\cite{clva}. (g)~AIF-B~\cite{aif}. (h)~AIF-D.}
  \label{fig:comparison}
\end{figure*}

\section{Experiments}
\label{sec:experiment}

\subsection{Quantitative evaluation metrics}
To conduct a thorough evaluation of the model's performance in the AIF task, we utilize four quantitative metrics: 
\textit{(i)} \textbf{Structural Similarity Index Measure (SSIM).}
Following CLVA~\cite{clva}, we use SSIM~\cite{ssim} to assess whether the synthesized images have similar visual content to the content images, evaluating whether the synthesized results preserve the content consistency.
\textit{(ii)} \textbf{Shallow Style Difference (SSD).}
Since the shallow layers of the VGG network extract color and texture features that are more aligned with AIF objectives, we calculate the style difference~\cite{deng2022stytr2} between these layers to measure the statistical distribution of visual features between synthesized results and anchor images. 
\textit{(iii)} \textbf{Sentiment Gap (SG).}
Following the preliminary work~\cite{aif}, we extract sentiment vectors of synthesized results and anchor images according to Yang~\etal~\cite{yang2018retrieving}. Their Euclidean distance is further calculated to determine whether synthesized images could evoke specific emotional responses.
\textit{(iv)} \textbf{Ensemble Accuracy (EAcc).}
We extend the Accuracy metric in our preliminary work~\cite{aif}, separately estimating emotional distributions from color, texture, style, and patch features. By ensembling these estimation results, we comprehensively evaluate whether the image filter reflects the intended emotions.

\subsection{Comparison with state-of-the-art methods}
As the AIF task is a recently proposed task with limited closely related work, we compare our results with relevant text-driven image editing methods (\ie, ManiGAN~\cite{manigan} and DiffusionCLIP~\cite{diffusionclip}) and style transfer methods (\ie, CLIPstyler~\cite{clipstyler} and CLVA~\cite{clva}).

\begin{figure*}[t]
   \centering
  \includegraphics[width=\linewidth]{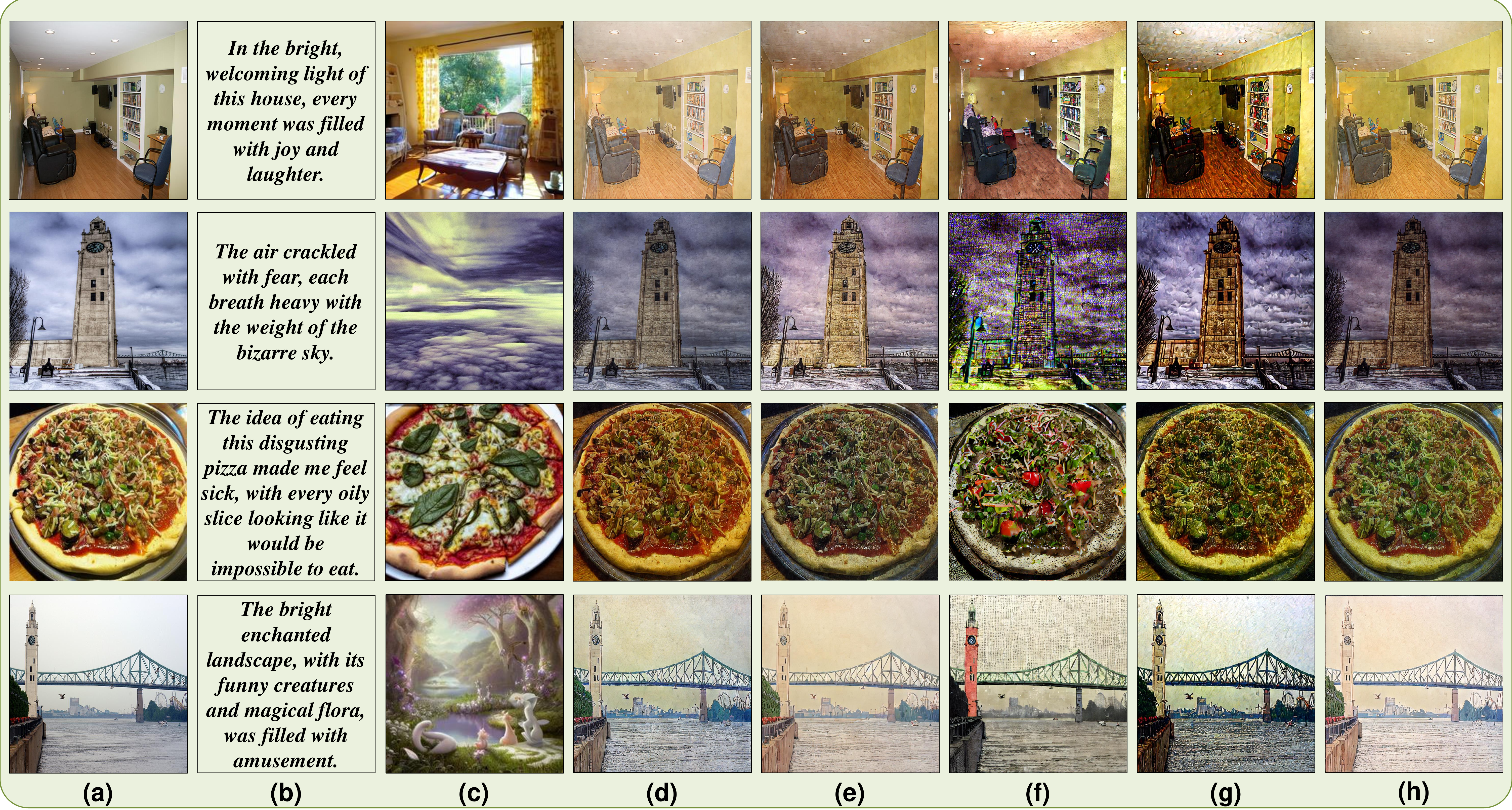}
  \caption{Ablation study results with different variants of AIF-D. (a)~User-provided content images. (b)~Text descriptions that reflect thoughts and feelings. (c)~W/o CPM. (d)~W/o IER. (e)~W/o VEM. (f)~W/o ERD. (g)~W/o TML. (h)~AIF-D. }
  \vspace{-2mm}
  \label{fig:ablation}
\end{figure*}

\begin{table}[t]
\centering
\caption{Quantitative experiment results. $\uparrow$ ($\downarrow$) means higher (lower) is better. Throughout the paper, best performances are highlighted in \textbf{bold}.}\label{tab:comparison}
\setlength\tabcolsep{5pt} 
\begin{adjustbox}{width={0.48\textwidth},totalheight={\textheight},keepaspectratio}
    \begin{tabular}{l|c c c c}
    \toprule
    Methods  &SSIM (\%) $\uparrow$ & SSD $\downarrow$  & SG (\textperthousand) $\downarrow$   &  EAcc (\%) $\uparrow$ \\ \midrule
    \multicolumn{5}{c}{Comparison with state-of-the-art methods} \cr \midrule
    ManiGAN & 50.72 & 1.4970 & 1.6589 & 33.58 \\ 
    DiffusionCLIP & 53.05 & 1.9873 & 1.7095 & 32.71  \\ 
    CLIPstyler & 52.49 & 1.9717 & 1.5676 & 34.19  \\ 
    CLVA & 50.30 & 1.2232 & 1.6707 & 32.32 \\ 
    \midrule
    AIF-B (Ours) & 56.15 & 1.1739 & 1.3881 & 36.02 \\
    AIF-D (Ours)  & \textbf{57.74} & \textbf{1.1123} & \textbf{1.3416} & \textbf{37.48} \\ \midrule \midrule
    
    \multicolumn{5}{c}{Ablation study of AIF-D} \cr \midrule
    \textit{W/o} CPM & 13.48 & 2.5374 & 1.8421 & 31.19 \\
    \textit{W/o} IER & 57.54 & 1.1251 & 1.3662 & 35.87 \\
    \textit{W/o} VEM & 57.32 & 1.1173 & 1.3949 & 34.18 \\
    \textit{W/o} ERD & 55.75 & 1.2356 & 1.4639 & 35.21 \\
    \textit{W/o} TML & 56.91 & 1.2824 & 1.4312 & 36.23 \\ 
    
    \bottomrule
    \end{tabular}
\end{adjustbox}
\vspace{-2mm}
\end{table}

\begin{table}[t]
\centering
\caption{User study results. AIF-D outperforms relevant methods with the highest score.}\label{tab:user_study}
\setlength\tabcolsep{11pt} 
\scriptsize 
\begin{adjustbox}{width={0.48\textwidth},totalheight={\textheight},keepaspectratio}
    \begin{tabular}{l|c c c}
    \toprule
    Methods  & EPS (\%) $\uparrow$ & EFS $\uparrow$ & FES (\%) $\uparrow$ \\ \midrule
    ManiGAN & $7.63$ & $3.01$ & $13.24$  \cr
    DiffusionCLIP & $11.72$ & $3.35$ & $11.88$ \cr
    CLIPstyler & $12.84$ & $3.37$ & $6.19$ \cr
    CLVA & $9.78$ & $3.03$ & $9.30$ \cr \midrule\midrule
    AIF-B (Ours) & $18.52$ & $3.45$ & $16.44$ \cr
    AIF-D (Ours) & $\textbf{39.51}$ & $\textbf{3.66}$ & $\textbf{42.95}$ \cr
    \bottomrule
    \end{tabular}
\end{adjustbox}
\vspace{-2mm}
\end{table}

\noindent \textbf{Qualitative comparisons.}
We show visual quality comparisons with the aforementioned methods in~\cref{fig:comparison}. 
Among these methods, ManiGAN~\cite{manigan} tends to distort content details (\eg, the first row, the flowers in the vase become blurred); 
DiffusionCLIP~\cite{diffusionclip} fails to preserve the overall structure of the content (\eg, the second row, the whole photo seems gritty);
CLIPstyler~\cite{clipstyler} tends to over-stylize the content (the third row, the filter creates an unnaturally radiant golden glow);
CLVA~\cite{clva} generally synthesizes colorless images, making it difficult to accurately convey emotions (\eg, the fourth row, the castle appears shrouded in a layer of gray haze); 
AIF-B~\cite{aif} can partially reflect emotions from text descriptions but struggles with effectively balancing artistic style and content consistency, resulting in blurry details (\eg, the fourth row, the texture on the castle walls looks irregular and distorted).  
In contrast, AIF-D demonstrates superior content consistency and emotional fidelity, advancing AIF models towards deeper emotional reflection.

\begin{table}[t]
\centering
\caption{Hyper-parameter sensitivity analysis for AIF-D model.}\label{tab:sensitivity}
\setlength\tabcolsep{5pt} 
\begin{adjustbox}{width={0.48\textwidth},totalheight={\textheight},keepaspectratio}
    \begin{tabular}{l|c c c c}
    \toprule
    Methods  &SSIM (\%) $\uparrow$ & SSD $\downarrow$  & SG (\textperthousand) $\downarrow$   &  EAcc (\%) $\uparrow$ \\ \midrule
    AIF-D ($\gamma = 0.1$) & 57.70 & 1.2057 & 1.4575 & 36.64 \\
    AIF-D ($\gamma = 1.0$) & 57.03 & 1.1679 & 1.3694 & 37.04 \\
    AIF-D ($\gamma = 3.0$) & 55.11 & 1.3142 & 1.6127 & 36.02 \\ \midrule\midrule
    AIF-D ($\gamma = 0.3$) & \textbf{57.74} & \textbf{1.1123} & \textbf{1.3416} & \textbf{37.48} \\

    \bottomrule
    \end{tabular}
\end{adjustbox}

\end{table}

\noindent \textbf{Quantitative comparisons.}
We show comprehensive quantitative results in~\cref{tab:comparison}, where AIF-D achieves the highest scores across all four metrics. This demonstrates that AIF-D outperforms state-of-the-art methods. Specifically, AIF-D effectively preserves the overall structure of the content image (SSIM), exhibits appropriate filter-like visual effects (SSD), evokes specific emotions of human observers (SG), and reflects emotions to images more accurately (EAcc).

\begin{figure*}[t]
   \centering
  \includegraphics[width=\linewidth]{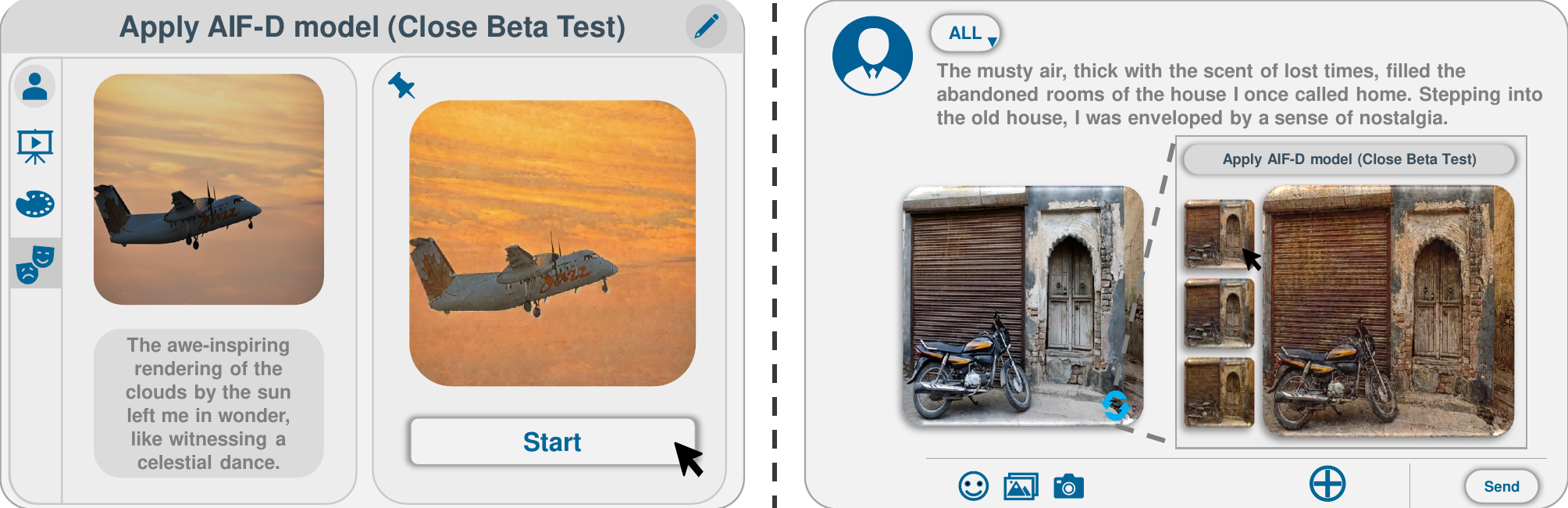}
  \caption{Application. 
  \textbf{Left:} Retouching photography. By leveraging text descriptions that reflect users' thoughts when taking the photo, AIF models can retouch these images to convey their intended emotions.
  \textbf{Right:} Social platforms. When sharing experiences on social platforms, users can select results that accurately reflect their emotions, thereby personalizing their content and attracting more followers.
  } 
  \label{fig:application}
  \vspace{-1mm}
\end{figure*}

\noindent \textbf{User study.}
In addition to qualitative and quantitative comparisons, we further conduct three user studies to evaluate the effectiveness of AIF models. 
\textit{(i)} \textbf{Emotional Preference Scores (EPS).}
We conduct a user study to determine whether images synthesized by AIF model are more favored by human observers compared to those produced by relevant state-of-the-art methods. In this study, each participant is shown a combination of a content image, an emotional text, and six synthesized images. They are asked to select the synthesized image that they feel best matches the emotional text descriptions. We denote the rate of selection as the EPS for each method.
\textit{(ii)} \textbf{Emotional Fidelity Scores (EFS).}
We investigate the emotional shift from the content images to the synthesized ones by asking participants to categorize the emotions of each synthesized image. The EFS ranging from 5 to 1 are calculated according to the emotional distance between the content images and synthesized images in Mikel's wheel~\cite{mikels2005emotional}, where higher scores mean closer categories with respect to the labels of corresponding text descriptions.
\textit{(iii)} \textbf{Filter-like Effect Scores (FES).}
To further evaluate the visual effects of AIF models, we ask users to judge which image has the strongest filter-like effect. Participants are shown six synthesized images from comparison methods and are asked to select the one that appears most like the result of applying a filter. The FES for each method represents the selection rate.
These experiments are conducted on Amazon Mechanical Turk (AMT), where 200 samples from the testing set of the AIF dataset are randomly selected, and experiment results are polled independently by 50 volunteers. We present the EPS, EFS, and FES results in Tab.~\ref{tab:user_study}, where the highest scores indicate the subjective advantages of AIF-D.
To further measure inter-rater agreement, we calculate Fleiss's Kappa~\cite{fleiss1971}, resulting in 0.60 and 0.63 for EPS and FES, indicating a substantial level of consistency among raters.

\begin{figure*}[t]
    \centering
  \includegraphics[width=\linewidth]{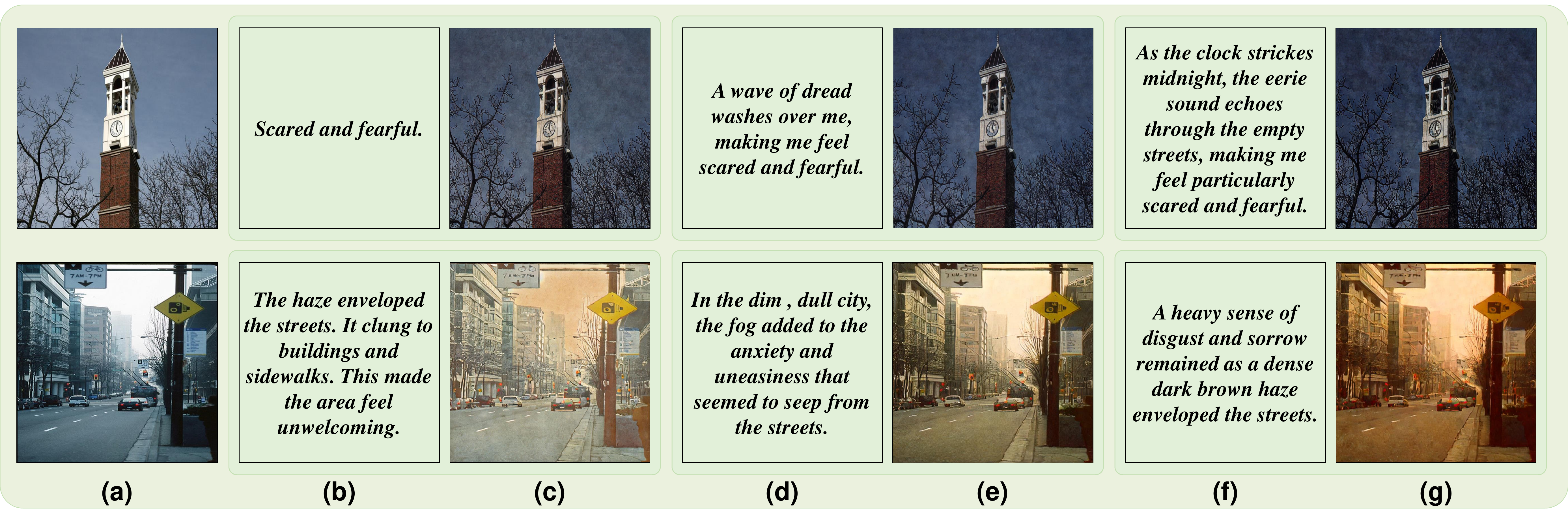}
  \caption{Robustness testing. \textbf{Top:}  Examples with varying levels of descriptive detail. \textbf{Bottom:} Examples with varying levels of emotional intensity. (a)~User-provided content images. (b, d, f)~Text descriptions that reflect thoughts and feelings. (c, e, g)~Robust results produced by AIF-D.}
  \label{fig:robust}
  \vspace{-2mm}
\end{figure*}

\subsection{Ablation study}
We discard various modules and create five variants to study the impacts of AIF-D's modules and designed losses. The evaluation scores and synthesized images of the ablation study are shown in~\cref{tab:comparison} and~\cref{fig:ablation}, respectively.

\noindent \textbf{W/o CPM (Content Preservation Module).} In this ablation, we remove the content preservation module. 
As a result, this ablation cannot preserve the overall structure without corresponding content images as cues (lower SSIM score). As shown in the first row of~\cref{fig:ablation}, synthesized images are completely irrelevant to the user-provided content images and are therefore unqualified.

\noindent \textbf{W/o IER (In-depth Emotional Reasoning).} 
We discard the in-depth emotional reasoning module. 
Due to the absence of emotional and filter effect prompts, AIF-D cannot leverage the emotional cues from the LLM to infer deeper emotions (higher SG score and lower EAcc score). As shown in the second row of~\cref{fig:ablation}, the overall color tone of the image appears dull, failing to convey the emotion of fear.

\noindent \textbf{W/o VEM (Voting Ensemble Mechanism).} 
We remove the voting ensemble mechanism. Consequently, AIF-D has difficulty in comprehensively understanding ambiguous and subjective emotions (reduced EAcc score). 
As shown in the third row of~\cref{fig:ablation}, the overall color tone of the image is on the darker side, failing to convey the disgusting feeling mentioned in the text description effectively.

\noindent \textbf{W/o ERD (Emotional Reflection Decoder).} 
We replace the emotional reflection image decoder with the vanilla image decoder. 
Therefore, this variant struggles to anchor emotions to colors and textures with appropriate artistic representations (higher SG score).
As shown in the fourth row of~\cref{fig:ablation}, there are numerous white and black artifacts scattered across the image, damaging the image emotional representation.

\noindent \textbf{W/o TML (Texture Mapping Loss).} 
During the training of the noise prediction network, we utilize the standard style loss~\cite{deng2022stytr2} instead of the proposed texture mapping loss. As a result, the model excessively applies artistic modifications to user-provided content images (higher SSD score). As shown in the second row of~\cref{fig:ablation}, the room presents exaggerated contrast, leading to overly-stylized effects.

\subsection{Hyper-parameter sensitivity analysis.}

We further investigate the influence of the crucial hyper-parameter $\gamma$ (introduced in~\cref{eq:tm}), which controls the balance between artistic style and content consistency. As~\cref{tab:sensitivity} demonstrates that $\gamma=0.3$ achieves the best quantitative results, we adopt this value for the AIF-D model.

\begin{figure*}[t]
    \centering
  \includegraphics[width=\linewidth]{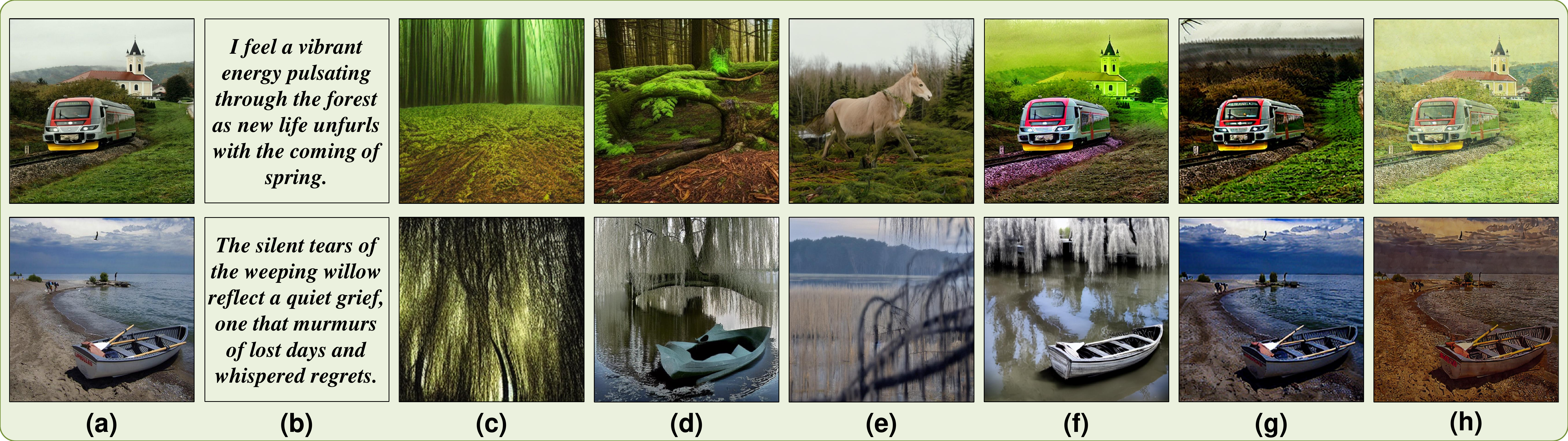}
  \caption{Qualitative comparison results with general image editing models based on the diffusion approach. (a) User-provided content images. (b) Texts that reflect thoughts and feelings. (c)  Stable Diffusion~\cite{stablediffusion}. (d) ControlNet~\cite{controlnet}. (e) SDEdit~\cite{SDEdit}. (f) InstructPix2Pix~\cite{instructpix2pix}. (g) Imagic~\cite{Imagic}. (h) AIF-D.}
  \label{fig:discussion}
\end{figure*}

\section{Discussion} \label{sec:discussion}

\subsection{Application}
The AIF models are typically applied in two scenarios.

\noindent \textbf{Retouching photography.} With daily photography becoming a ubiquitous practice, there is a growing expectation for photos to have artistic qualities, be visually pleasing, and reflect the photographer's mood. Given users' written thoughts, AIF models could retouch their photos to be visually compelling and reflect what they are thinking. We show this application in~\cref{fig:application} (left), where the user requests the model to enhance their photograph by applying affective filters that reflect the emotion they experienced at the moment of capture.

\noindent \textbf{Social platforms.} People share posts with texts and corresponding images to express their feelings on social platforms. AIF models could personalize their content, potentially enabling influencers to stand out from the crowd and evoke specific emotional responses from their followers. As demonstrated in~\cref{fig:application} (right),  several affective images are provided as options, recognizing that emotions are inherently subjective and ambiguous. This allows users to select the result that most accurately represents their intended visually-abstract emotions.

\subsection{Robustness testing of AIF-D}
To evaluate the robustness of AIF-D's text understanding with diverse user input (\eg, varying levels of details and emotional intensity), we conduct the following experiments.

\noindent \textbf{Levels of detail.} As shown in~\cref{fig:robust} (top), we present examples with consistent emotional intensity but varying levels of detail. As a result, AIF-D effectively conveys emotions (particularly scared and fearful), regardless of the levels of descriptive detail.

\noindent \textbf{Levels of intensity.} As shown in \cref{fig:robust} (bottom), we maintain descriptive detail but change emotional intensity. Consequently, AIF-D successfully produces results that progressively heighten the sense of emotions (disgust and discomfort).

\subsection{Differences from general image editing}
Reflecting visually-abstract emotions from text descriptions to visually concrete images presents a significant challenge. 
Existing general image editing techniques achieve impressive results in their respective domains but struggle to meet the objectives of the AIF task.
Specifically, Stable Diffusion~\cite{stablediffusion} and SDEdit~\cite{SDEdit} struggle to preserve the overall structure of content images; ControlNet~\cite{controlnet}, InstructPix2Pix~\cite{instructpix2pix}, and Imagic~\cite{Imagic} are unable to understand emotional cues to synthesize appropriate artistic representation. 
We illustrate their unsuitability in~\cref{fig:discussion}.

\section{Conclusion}
\label{sec:conclusion}
In this paper, we introduce the task of Affective Image Filter (AIF), which requires the model to understand visually-abstract emotions from text descriptions and reflect them into visually-concrete images with appropriate colors and textures.
We first introduce the AIF dataset and formulate the AIF model. 
Next, we present AIF-B, built upon a multi-modal transformer architecture.
As an initial attempt, AIF-B leverages external emotional cues, proposing the emotional reflection loss, and designing the visual aesthetic loss.
To advance AIF models towards deeper emotional reflection, we propose AIF-D, a model built upon a pre-trained diffusion model with a content preservation module for preserving high-frequency detail. 
By ensembling emotional cues, improving emotional reflection strategy, and redesigning visual aesthetic balance, AIF-D achieves state-of-the-art performance across four quantitative and three user study experiments.
Finally, we discuss applications, robustness, and differences from general image editing techniques to highlight the value and potential of AIF models.

\section*{Acknowledgement}
This work is supported by National Natural Science Foundation of China (Grant No. 62136001) and Beijing Municipal Science \& Technology Commission, Administrative Commission of Zhongguancun Science Park (Grant No. Z241100003524012). 
We thank vivo for providing the GPU resources and thank Qingnan Fan and Jinwei Chen for their insightful discussions and valuable feedback.

\bibliographystyle{IEEEtran}
\bibliography{main}

\vspace{25mm}

\end{document}